\documentclass{template/llncs}

\usepackage{template/llncsdoc}

\usepackage{graphicx}
\usepackage{amssymb,amsmath,bm}
\usepackage{textcomp}
\usepackage{url}
\usepackage{framed}
\usepackage[nolist]{acronym}

\begin{acronym}
    \acro{ASR}{Automatic Speech Recognition}
    \acro{CMU}{Carnegie Mellon University}
    \acro{FCT}{Funda\c{c}\~{a}o para a Ci\^{e}ncia e a Tecnologia}
    \acro{IVR}{Interactive Voice Response}
    \acro{NLP}{Natural Language Processing}
    \acro{WER}{Word Error Rate}
\end{acronym}

\begin{document}

\thispagestyle{empty}

\title{Mapping the Dialog Act Annotations of the LEGO Corpus into the Communicative Functions of ISO 24617-2}
\author{Eug\'{e}nio Ribeiro\inst{1,2} \and Ricardo Ribeiro\inst{1,3} \and David Martins de Matos\inst{1,2}}
\institute{L$^2$F {--} Spoken Language Systems Laboratory {--} INESC-ID Lisboa
\and
Instituto Superior T\'{e}cnico, Universidade de Lisboa, Portugal
\and
ISCTE-IUL {--} Instituto Universit\'{a}rio de Lisboa, Portugal
\email{eugenio.ribeiro@l2f.inesc-id.pt}}

\maketitle

%
%

\begin{abstract}

In this paper we present strategies for mapping the dialog act annotations of the LEGO corpus into the communicative functions of the ISO 24617-2 standard. Using these strategies, we obtained an additional 347 dialogs annotated according to the standard. This is particularly important given the reduced amount of existing data in those conditions due to the recency of the standard. Furthermore, these are dialogs from a widely explored corpus for dialog related tasks. However, its dialog annotations have been neglected due to their high domain-dependency, which renders them unuseful outside the context of the corpus. Thus, through our mapping process, we both obtain more data annotated according to a recent standard and provide useful dialog act annotations for a widely explored corpus in the context of dialog research.     

\end{abstract}

%
%

%
%

\section{Introduction}
\label{sec:introduction}

During a conversation, dialog acts are the minimal units of linguistic communication, since they are able to reveal the intention behind the uttered words~\cite{Searle1969}. Thus, they play an important role and have been widely explored by linguists and computer scientists in the \ac{NLP} area. In this sense, automatic dialog act recognition is particularly important in the context of dialog systems~\cite{Kral2010}. However, in order to develop systems that are able to successfully recognize dialog acts, annotated data is required. This takes us to the problem of dialog act annotation and the multiple methodologies and tag sets described in the different studies performed along the years. Dialog act annotation was typically performed in the context of projects or the development of datasets, each introducing new tag sets or modifying the existing ones. This led to a wide scattering of data in terms of the used annotation, which hardens the comparison of results and conclusions obtained using different approaches. Thus, in an attempt to set the ground for more comparable research in the area, a standard, ISO 24617-2~\cite{Bunt2012}, was developed. However, since the standard is relatively recent and the described annotation process is non-trivial, the amount of data annotated according to it is reduced. In order to increase that amount, we defined strategies for mapping the original dialog act annotations of the LEGO corpus~\cite{Schmitt2012} into the communicative functions of the standard. We chose this corpus because although it has been widely explored in dialog related tasks, its original dialog act annotations have been neglected. This is due to the high domain dependence of the labels, which can only be used on that specific dataset. However, such specificity also simplifies the mapping of the labels into the higher level domain-independent communicative functions of the standard.

In the remaining sections of this paper we start by providing some insight into the ISO 24617-2 and the LEGO corpus in Sections~\ref{sec:iso} and~\ref{sec:lego}, respectively. After that, in Sections~\ref{sec:system} and~\ref{sec:user}, we describe our mapping strategies for both system and user turns of the corpus. Finally, in Section~\ref{sec:discussion}, we discuss the distribution of the communicative functions of the standard in the corpus and the quality of the conversion process.            

%
%

%
%

\section{ISO 24617-2}
\label{sec:iso}

\begin{figure}[htbp]
    \includegraphics[width=\textwidth]{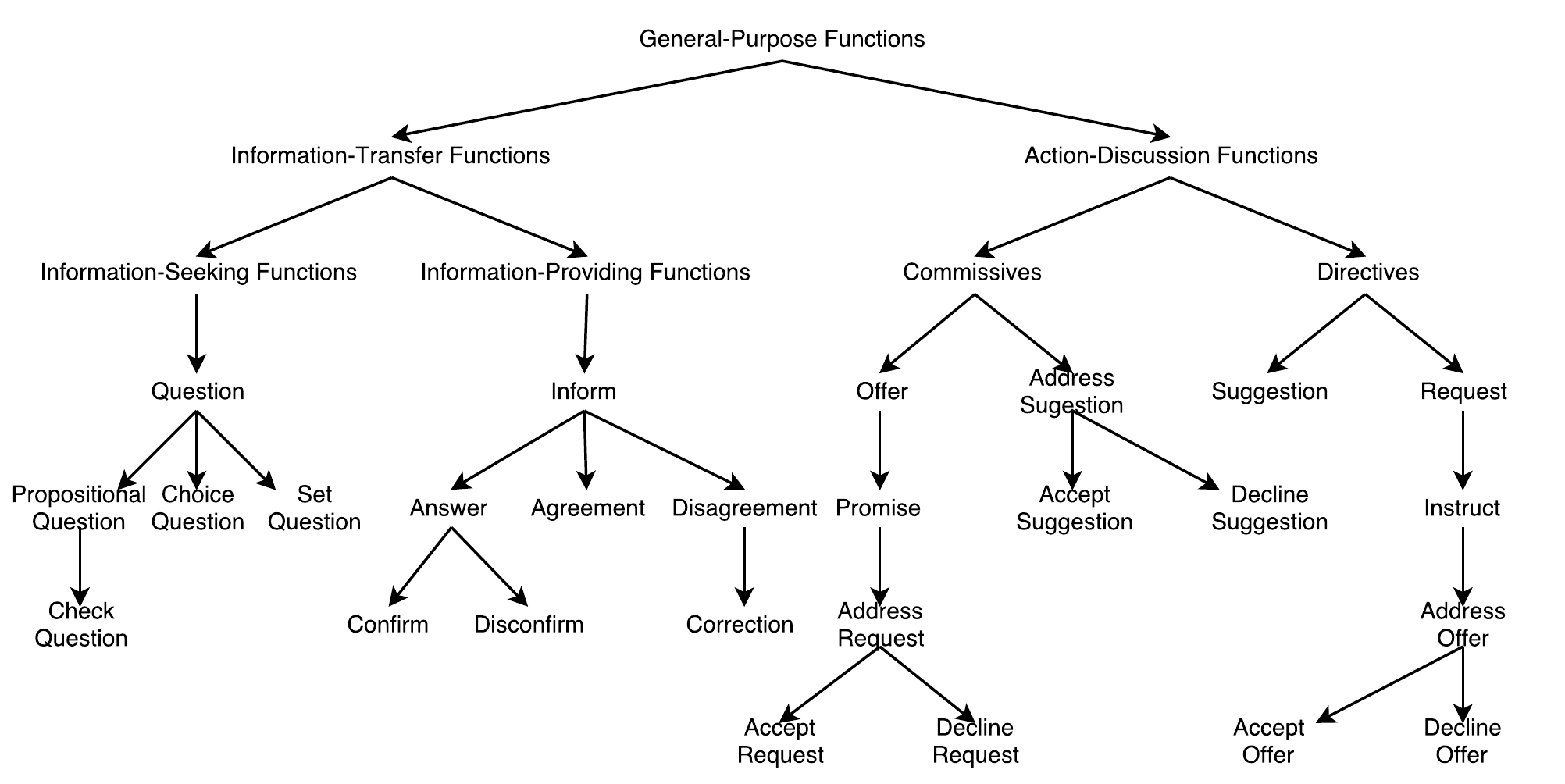}
    \caption{Distribution of general-purpose communicative functions according to the ISO 24617-2 standard. Adapted from~\protect\cite{Bunt2010}.}
    \label{fig:generaldistr}
\end{figure}

\begin{figure}[htbp]
    \includegraphics[width=\textwidth]{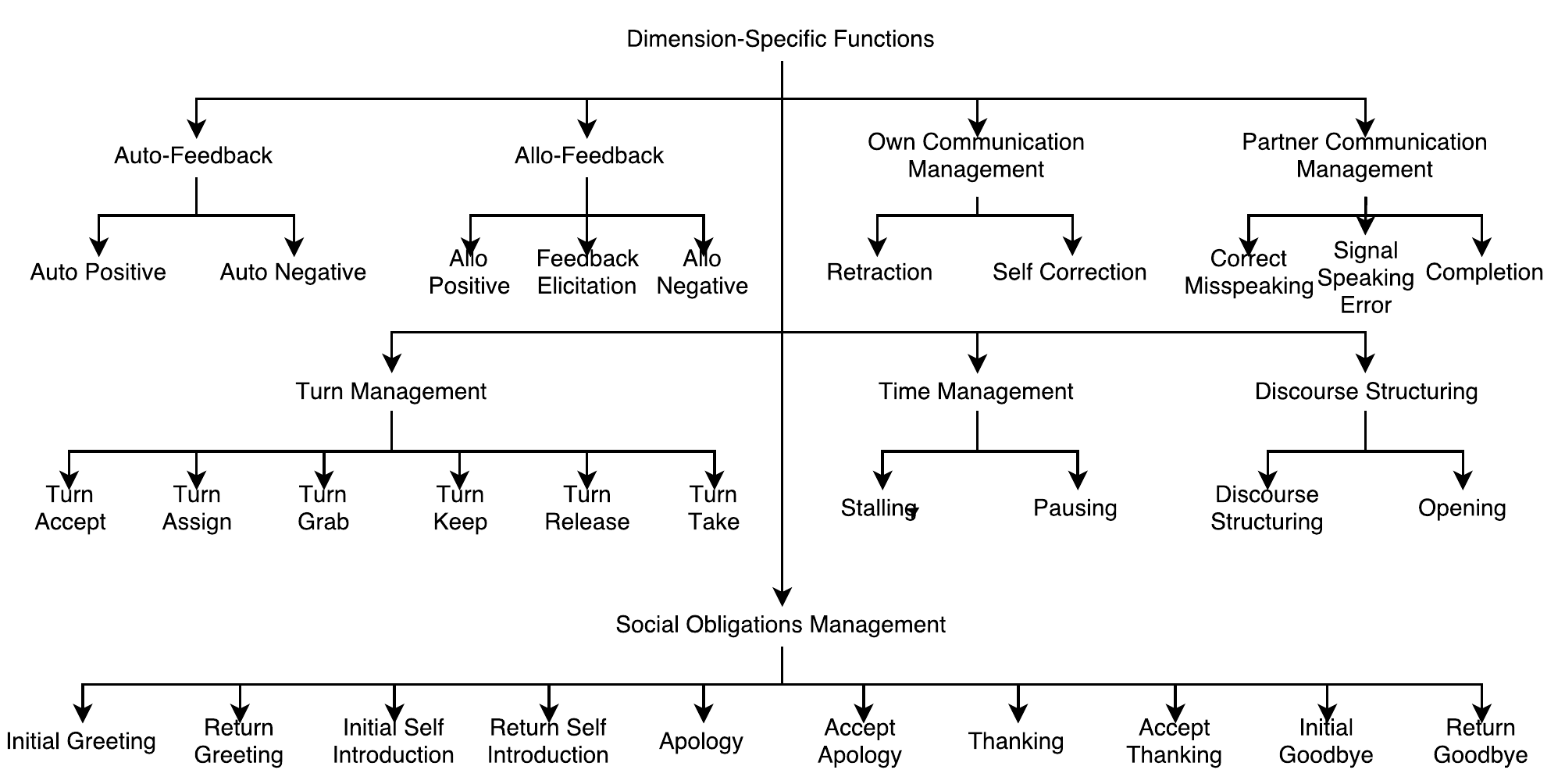}
    \caption{Distribution of dimension-specific communicative functions according to the ISO 24617-2 standard.}
    \label{fig:dimensiondistr}
\end{figure}

In an attempt to standardize dialog act annotation and, thus, set the ground for more comparable research in the area, Bunt et al.~\cite{Bunt2012} defined the ISO 24617-2 standard. The first thing that should be noticed in the standard is that annotations should be performed on functional segments rather than on turns or utterances~\cite{Carroll1978}. This should happen because a single turn or utterance may have multiple functions, revealing different intentions. However, automatic, and even manual, functional segmentation is a complex task on its own. Furthermore, according to the standard, dialog act annotation does not consist of a single label, but rather of a complex structure containing information about the participants, relations with other functional segments, the semantic dimension of the dialog act, its communicative function, and optional qualifiers concerning certainty, conditionality, partiality, and sentiment. In terms of semantic dimensions, the standard defines nine {--} \em Task\em, \em Auto-Feedback\em, \em Allo-Feedback\em, \em Turn Management\em, \em Time Management\em, \em Discourse Structuring\em, \em Own Communication Management\em, \em Partner Communication Management\em, and \em Social Obligations Management\em. Communicative functions correspond to the dialog act labels present in the multiple tag sets used to annotate data before the introduction of the standard~\cite{Allen1994,Carletta1996,Jurafsky1997,Allen1997,DiEugenio1998,Alexandersson1998}. They were divided into general-purpose functions, which can occur in any semantic dimension, and dimension-specific functions, which, as the name indicates, are specific to a certain dimension. The set of general-purpose functions is hierarchically distributed according to Figure~\ref{fig:generaldistr}. Dimension-specific functions are all at the same level and are distributed across dimensions according to Figure~\ref{fig:dimensiondistr}. We can see that there are dimension-specific functions for only eight of the nine dimensions. This means that the task dimension contains general-purpose functions only. Furthermore, although there are ten communicative functions specific for the Social Obligations Management dimension, some consist of the initial and return counterparts of the same function. These are usually paired to form higher level communicative functions.

Since the standard is relatively recent and the annotation process is non-trivial, the amount of data annotated according to it is reduced. In this sense, the largest amount of annotated data is provided by the Tilburg University DialogBank~\cite{Bunt2016}~\footnote{\url{https://dialogbank.uvt.nl/}}. If features dialogs from different corpora in multiple languages. Most of the annotated data was obtained through conversion processes of previously annotated dialogs from corpora such as Switchboard~\cite{Godfrey1992} and HCRC Map Task~\cite{Thompson1993}. However, in order to obtain a complete annotation, these conversion processes, such as the one for the Switchboard corpus~\cite{Fang2012}, require manual steps, which are both time and resource consuming. This is especially notorious for resegmentation processes of turns and utterances into functional segments. Thus, in spite of the efforts, some of the dialogs in the DialogBank are not completely annotated according to the standard and the addition of new dialogs is slow. Overall, this means that in order to disseminate the standard and perform solid experiments using its annotation methodology, it is important to put some effort into obtaining larger amounts of data annotated according to it.

%
%

%
%

\section{LEGO Corpus}
\label{sec:lego}

The LEGO corpus~\cite{Schmitt2012} is an annotated subset of the The Let's Go corpus~\cite{Raux2006}. It features data from the \ac{CMU}'s Let's Go Bus Information System, which provides information about bus schedules in the city of Pittsburg, through spoken telephonic interaction with a dialog system.

A set of 347 calls recorded during 2006 was later annotated by the Dialogue Systems Group at Ulm University, Germany. This annotated subset is the one usually referred to as LEGO. It features 14186 utterances {--} 9083 system utterances and 5103 user utterances. The number of utterances is different because some system turns contain multiple utterances and some of its prompts are not answered by the user. There is no standard division of this corpus into training and testing sets.

Since system utterances are generated through slot filling of fixed templates, they have no errors and contain casing and punctuation information. In contrast, the transcriptions of user utterances were obtained using an \ac{ASR} system and, thus, contain no casing nor punctuation information. Furthermore, the recognition was not always correct, due to the error rate of the system. Unfortunately, a concrete value for the \ac{WER} is not revealed. Figure~\ref{fig:legodialog} presents an excerpt of a transcription. 

\begin{figure}[htbp]
    \begin{framed}
        \textbf{System:} Welcome to the CMU Let's Go bus information system.
        
        \textbf{System:} To get help at any time, just say Help or press zero.
        
        \textbf{System:} What can I do for you?
        
        \textbf{User:} PENN AT BUTLER TIME IS IT IS THE CENTRE AT HIGHLAND FROM RUNS THE 61A EIGHT THIRTY 61B CAN'T OAKLAND
        
        \textbf{System:} Leaving from OAKLAND . Is this correct?
        
        \textbf{User:} OAKLAND
        
        \textbf{System:} Leaving from OAKLAND . Is this correct?
        
        \textbf{User:} YES
        
        \textbf{System:} Right.
        
        \textbf{System:} Where do you want to go?
        
        \textbf{User:} SQUIRREL HILL OF FIFTH AVENUE
        
        \textbf{System:} Going to FIFTH AVENUE . Is this correct?
        
        \textbf{User:} YES
    \end{framed}
    \caption{An excerpt of a LEGO corpus transcription.}
    \label{fig:legodialog}
\end{figure}

In terms of dialog acts, the original LEGO corpus annotations relied on two distinct and domain-dependent tag sets for system an user turns. The set for system turns contains 28 labels, such as \em Ask Destination\em, \em Ask Bus\em, and \em Ask Time\em, while the set for user turns contains 22 labels, such as \em Place Information\em, \em Confirm Destination\em, and \em Reject Bus\em. When using such tag sets, context information is clearly very important both for dialog act annotation and recognition, since a given dialog act drastically reduces the number of non-disruptive possibilities, that is, that do not break the dialog flow, for the next one. Tables~\ref{tab:systemdistr} and~\ref{tab:userdistr} show the distribution of the labels among the system and user turns, respectively. It is important to notice that the second most frequent label in user turns is \em Unqualified / Unrecognized\em, accounting for 15.09\% of those turns, which reveals a high number of communication problems between the user and the system. 

\begin{table}[htbp]
    \caption{Original label distribution in the system turns of the LEGO corpus.}
    \label{tab:systemdistr}
    \centering
    \begin{tabular}{l r r}
        \hline
        \textbf{Label}           & \textbf{Count} & \textbf{\%} \tabularnewline
        \hline
        Confirm Understood       &   912     &  10.04 \tabularnewline
        Confirm Departure        &   823     &   9.06 \tabularnewline
        Ask Another Query        &   728     &   8.01 \tabularnewline
        Ask Bus                  &   714     &   7.86 \tabularnewline
        Ask Confirm Time         &   497     &   5.47 \tabularnewline
        Ask Time                 &   493     &   5.43 \tabularnewline
        Ask Confirm Destination  &   458     &   5.04 \tabularnewline
        Explain                  &   438     &   4.82 \tabularnewline
        Ask Confirm Bus          &   425     &   4.68 \tabularnewline
        Ask Departure            &   410     &   4.51 \tabularnewline
        Deliver Result           &   410     &   4.51 \tabularnewline
        Filler                   &   410     &   4.51 \tabularnewline
        Announce Querying        &   364     &   4.01 \tabularnewline
        Offer Help               &   348     &   3.83 \tabularnewline
        Greeting                 &   347     &   3.82 \tabularnewline
        Ask Destination          &   345     &   3.80 \tabularnewline
        Inform No Schedule       &   343     &   3.78 \tabularnewline
        Ask Confirm With Keys    &   142     &   1.56 \tabularnewline
        Ask Confirm Neighborhood &   128     &   1.41 \tabularnewline
        Announce Restart         &    84     &   0.92 \tabularnewline
        Inform Shorter Answer    &    75     &   0.82 \tabularnewline
        Inform Help              &    67     &   0.74 \tabularnewline
        Goodbye                  &    36     &   0.40 \tabularnewline
        Disambiguate Bus Stop    &    32     &   0.35 \tabularnewline
        Inform No Route          &    25     &   0.28 \tabularnewline
        Instruct Louder          &    13     &   0.11 \tabularnewline
        Confirm Restart          &    10     &   0.07 \tabularnewline
        Instruct More Quiet      &     6     &   0.01 \tabularnewline
    \end{tabular}
\end{table}

\begin{table}[htbp]
    \caption{Original label distribution in the user turns of the LEGO corpus.}
    \label{tab:userdistr}
    \centering
    \begin{tabular}{l r r}
        \hline
        \textbf{Label}              & \textbf{Count} & \textbf{\%} \tabularnewline
        \hline
        Place Information           &   879     &  16.94 \tabularnewline
        Unqualified / Unrecognized  &   783     &  15.09 \tabularnewline
        Reject                      &   746     &  14.38 \tabularnewline
        Line Information            &   440     &   8.48 \tabularnewline
        Time Information            &   391     &   7.54 \tabularnewline
        Confirm Departure           &   291     &   5.61 \tabularnewline
        Confirm Destination         &   246     &   4.74 \tabularnewline
        Confirm Time                &   225     &   4.34 \tabularnewline
        Confirm                     &   214     &   4.12 \tabularnewline
        Confirm Bus                 &   179     &   3.45 \tabularnewline
        Request Next Bus            &   159     &   3.06 \tabularnewline
        Reject Departure            &   135     &   2.60 \tabularnewline
        New Query                   &    98     &   1.89 \tabularnewline
        Reject Time                 &    95     &   1.83 \tabularnewline
        Request Previous Bus        &    75     &   1.44 \tabularnewline
        Reject Bus                  &    69     &   1.33 \tabularnewline
        Reject Destination          &    53     &   1.02 \tabularnewline
        Request Help                &    52     &   1.00 \tabularnewline
        Goodbye                     &    29     &   0.59 \tabularnewline
        Request Schedule            &    18     &   0.35 \tabularnewline
        Polite                      &     8     &   0.15 \tabularnewline
        Inform                      &     3     &   0.06 \tabularnewline
    \end{tabular}
\end{table}

Although the LEGO corpus has been used in many research tasks related to dialog and interaction with \ac{IVR} systems~\cite{Ultes2013,Sidorov2014,Brester2015,Griol2016}, its dialog act annotations have been neglected. In fact, to our knowledge, only we have used them in some dialog act recognition experiments in the context of the SpeDial project~\cite{Iosif2015}~\footnote{\url{http://www.spedial.eu}} probably due to the domain dependence of the labels, which would not be useful in any other domain. Furthermore, since the labels are so specific, even a system dealing with the same domain would only be able to benefit from them if the dialog had the same characteristics as the ones from the LEGO corpus. However, such specificity also leads to almost direct mappings of these labels into the higher level communicative functions of the ISO 24617-2 standard. This mapping does not produce a complete annotation according to the standard for three main reasons. First, not all the communicative functions present in the turns are covered, since it is not possible to obtain information related to certain dimensions from the transcriptions and the original labels alone. Secondly, communicative functions are just part of the annotation. Thirdly, according to the standard, annotations are made at the functional segment level and not at the turn level. Still, the mapping of the original labels into the communicative functions of the standard is able to provide a large amount of data for applications in the area of dialog act recognition. In fact, as stated in Section~\ref{sec:iso}, a large part of the existing data annotated according to the standard was annotated through conversion processes and does not provide all the required information to form a complete annotation according to the requirements of the standard. Thus, in the next sections we describe our strategies to map the original dialog act annotations of the LEGO corpus into the communicative functions of the ISO 24617-2 standard.

%
%

%
%

\section{Mapping of System Labels}
\label{sec:system}

Since the system turns of the LEGO corpus are generated through slot filling, each dialog act label is attributed to a small group of sentence templates. Thus, in order to map the labels into the communicative functions of the standard we analyzed those templates and attributed them the corresponding function. The strategies used for each label are presented below.    

\subsection{Announce Querying}
The turns annotated with this label consist of wait requests while the system looks for the required information, such as \em ``Just a second.'' \em and \em ``Hold on. I'll look that up.''\em. These are techniques for pausing the dialog which fall under the \em Time Management \em dimension defined in the standard and, thus, we annotated them with the \em Pausing \em communicative function. Furthermore, in cases such as the second example, the system makes a promise to look for the required information. Thus, those cases were also annotated with the \em Promise \em label in the \em Task \em dimension.

\subsection{Announce Restart}
All turns annotated with this label contain the same utterance {--} \em ``Okay, let's start from the beginning.'' \em {--} which reveals the intention of the system to structure the dialog towards a restart of the current interaction. Thus, we annotated these turns with the \em Interaction Structuring \em communicative function in the \em Discourse Structuring \em dimension.

\subsection{Ask Another Query}
The turns annotated with this label state the different options the user has at that point in an attempt to elicit one of the behaviors. Examples of such turns are \em ``You can say, when is the next bus, when is the previous bus, start a new query, or goodbye.'' \em and \em ``To ask about a different trip, you can say, start a new query. If you are finished, you can say goodbye.''\em. On the one hand, these turns provide instructions to the user about the task and, thus, were annotated with the \em Instruct \em communicative function in the \em Task \em dimension. On the other hand, they are also an attempt to structure the dialog and, thus, were also annotated with the \em Interaction Structuring \em communicative function in the \em Discourse Structuring \em dimension.

\subsection{Ask Bus}
\label{ssec:askbus}
In the cases annotated with this label, the system asks what it can do for the user, or what bus schedule he or she wants to obtain information on. Examples are \em ``What can I do for you?'' \em and \em ``What bus schedule information are you looking for?''\em. These are \em Set Questions \em in the \em Task \em dimension and, thus, were annotated accordingly. 

\subsection{Ask Departure}
\label{ssec:askdeparture}
These turns are similar to the ones annotated with the \em Ask Bus \em label (Section~\ref{ssec:askbus}), but asking for the departure place instead of a specific bus identifier. An example is \em ``Where are you leaving from?''\em. Accordingly, we annotated them with the \em Set Question \em communicative function in the \em Task \em dimension as well.

\subsection{Ask Destination}
These turns are in all similar to the ones annotated with the \em Ask Departure \em label (Section~\ref{ssec:askdeparture}), but asking for the destination place instead of the departure place. An example is \em ``What is you destination''\em . Thus, they were also annotated with the \em Set Question \em communicative function in the \em Task \em dimension.

\subsection{Ask Time}
Similarly to the previous three labels, turns annotated with this label consist of the system asking for information, in this case about the time of travel. An example is \em ``When do you wanna travel?''\em. Consequently, we also annotated them with \em Set Question \em communicative function in the \em Task \em dimension.

\subsection{Ask Confirm Bus}
The turns annotated with this label consist of two segments. The first states the identifier of the bus, as understood by the system, which is a case of auto-feedback, leading to annotation with the \em Auto Positive \em communicative function in the \em Auto-Feedback \em dimension. The second asks for confirmation by the user, which is a \em Check Question \em in the \em Task \em dimension. Examples of such turns are \em ``The 54C. Did I get that right?'' \em and \em ``The 28X. Is this correct?''\em.

\subsection{Ask Confirm Departure}
\label{ssec:askconfirmdeparture}
These turns are similar to the ones annotated with the previous label, but confirming information about the place of departure. An example is \em ``Leaving from Oakland. Is this correct?''\em . Thus, they were also annotated with the \em Auto Positive \em and \em Check Question \em communicative functions, in the \em Auto-Feedback \em and \em Task \em dimensions, respectively. However, some of the turns seem to be wrongly annotated, as they correspond to instructions by the system on how to confirm or reject using the keys and, thus, should have the \em Ask Confirm With Keys \em label. These cases were converted using the strategy for that label, as described in Section~\ref{ssec:keys}.

\subsection{Ask Confirm Destination}
These turns are in all similar to the ones annotated with the \em Ask Confirm Departure \em label (Section~\ref{ssec:askdeparture}), but concerning the destination place instead of the departure place. An example is \em ``Going to Fifth Avenue. Is this correct?''\em. Consequently, they were also annotated with the \em Auto Positive \em and \em Check Question \em communicative functions, in the \em Auto-Feedback \em and \em Task \em dimensions, respectively. Once again, some of the turns should have the \em Ask Confirm With Keys \em label instead. These cases were converted using the strategy for that label, as described in Section~\ref{ssec:keys}.

\subsection{Ask Confirm Neighborhood}
Turns annotated with this label are similar to the ones annotated with the previous two labels, but confirming a zone instead of a specific stop. An example is \em ``Waterworks Mall. Is this correct?''\em. Accordingly, they were also annotated with the \em Auto Positive \em and \em Check Question \em communicative functions, in the \em Auto-Feedback \em and \em Task \em dimensions, respectively.

\subsection{Ask Confirm Time}
Similarly to the previous four labels, turns annotated with this label consist of the system trying to confirm the the understood information, in this case concerning the time of travel. An example is \em ``Leaving at 5 a.m.. Did I get that right?''\em. Thus, we also annotated these turns with the \em Auto Positive \em and \em Check Question \em communicative functions, in the \em Auto-Feedback \em and \em Task \em dimensions, respectively. 

\subsection{Ask Confirm With Keys}
\label{ssec:keys}
Similarly to the previous labels, turns annotated with \em Ask Confirm With Keys \em states the understood information to provide feedback and requests confirmation. However, in this case, instead of making a question, the system instructs the user on how to answer using the keypad. An example is \em ``If you want the schedule of the 54C say yes or press one, otherwise say no or press three.''\em. Thus, consistently with the previous labels, we annotated these turns with the \em Auto Positive \em communicative function in the \em Auto-Feedback \em dimension. However, in the \em Task \em dimension, we annotated them with the \em Instruct \em communicative function instead of \em Check Question\em.

\subsection{Confirm Restart}
All turns annotated with this label contain the same utterance {--} \em ``Are you sure you want to start over?'' \em {--} which is a check by the system on whether the user really wants to restart the interaction. Thus, we annotated these turns with the \em Check Question \em communicative function in the \em Task \em dimension. Furthermore, since the system states that it understood a restart request, we also annotated these turns with the \em Auto Positive \em communicative function in the \em Auto-Feedback \em dimension.

\subsection{Confirm Understood}
\label{ssec:understood}
These turns signal the system's understanding of the user's intention, through feedback utterances such as \em ``Right'' \em and \em ``Ok''\em. Accordingly, we annotated them with the \em Auto Positive \em communicative function in the \em Auto-Feedback \em dimension.

\subsection{Deliver Result}
In the turns annotated with this label the system provides the bus schedule information according to the parameters discussed along the dialog. An example is \em ``The next 61C leaves Eighth Avenue at Ann at 7:45 p.m. and arrives at Second Street at Grant at 7:59 p.m..''\em. Thus, we annotated them with the \em Inform \em communicative function in the \em Task \em dimension.

\subsection{Disambiguate Bus Stop}
The turns annotated with this label occur when there is some confusion about the requested bus stops. Cases like \em ``Which stop in Duquesne are you leaving from?'' \em are questions that ask for a specific bus stop instead of a zone, while cases like \em ``Downtown and Forbes are both the same stop. Please provide a different start or end point.'' \em request a different start or ending point. Thus, in the \em Task \em dimension, we annotated turns similar to the first example with the \em Set Question \em communicative function and ones similar to the second example with \em Request \em.   

\subsection{Explain}
The turns annotated with this label provide instructions and examples to the user according to what he or she can do at that time. An example is \em ``For example, you can say, Forbes and Murray, Downtown, or McKeesport.''\em. Accordingly, we annotated these turns with the \em Instruct \em communicative function in the \em Task \em dimension.

\subsection{Filler}
All turns annotated with this label contain the same utterance {--} \em ``Alright'' \em {--} and serve the same purpose as the ones annotated as \em Confirm Understood \em (Section~\ref{ssec:understood}). Consequently, we also annotated them with the \em Auto Positive \em communicative function in the \em Auto-Feedback \em dimension.

\subsection{Goodbye}
\label{ssec:sdagoodbye}
These turns correspond to the system politely ending the dialog, with utterances such as \em ``Thank you for using the CMU Let's Go Bus Information System. Goodbye.'' \em. Thus, they fall on the \em Social Obligations Management \em dimension of the standard and were annotated with the \em Goodbye \em communicative function. 

\subsection{Greeting}
All turns annotated with this label contain the same utterance {--} \em ``Welcome to the CMU Let's Go bus information system.'' \em {--} which opens the interaction and greets the user. Thus, it has two functions on two different dimensions and, accordingly, we annotated them with the \em Opening \em communicative function in the \em Discourse Structuring \em dimension and with \em Greeting \em in the \em Social Obligations Management \em dimension.

\subsection{Inform Help}
All turns annotated with this label contain the same utterance {--} ``I am an automated spoken dialogue system that can give you schedule information for bus routes in Pittsburgh's East End. You can ask me about the following buses: 28X, 54C, 56U, 59U, 61A, 61B, 61C, 61D, 61F, 64A, 69A, and 501.'' {--} which informs the user of the buses that the system has information about. Thus, we annotated them with the \em Inform \em communicative function in the \em Task \em dimension.

\subsection{Inform No Route}
The turns annotated with this label state that there are no buses satisfying the indicated parameters and apologize for that. An example is \em ``I'm sorry, but there is no bus that goes between CMU and Squirrel Hill at that time.''\em. Thus, we annotated them with the \em Inform \em communicative function in the \em Task \em dimension. Furthermore, to cover the apologizing function, we also annotated them with the \em Apology \em communicative function in the \em Social Obligations Management \em dimension.  

\subsection{Inform No Schedule}
Similarly to the previous label, the turns annotated with this label state that the system does not have the schedule for the requested bus and apologize for that. An example is \em ``I'm sorry but I do not have the schedule for the 500. The routes I currently cover are the following: 28X, 54C, 56U, 59U, 61A, 61B, 61C, 61D, 61F, 64A, 69A, and 501.''\em. Thus, they were also annotated with the \em Inform \em and \em Apology \em communicative functions in the \em Task \em and \em Social Obligations Management \em dimensions, respectively. However, many turns were wrongly annotated with this label, which required manual and individual mapping.

\subsection{Inform Shorter Answer}
In these turns, the system asks the user to use shorter answers, both using polite requests, such as \em ``Please use shorter answers because I have trouble understanding long sentences.''\em, and more assertive commands, such as \em ``I need you to give me a short answer.''\em. In the \em Task \em dimension, the first were annotated with the \em Request \em communicative function, while the second were annotated with \em Instruct\em.

\subsection{Instruct Louder}
In these turns the system asks the user to speak louder using a polite request {--} \em ``I'm having some trouble hearing you. If you're still there, please try to talk a little bit louder or closer to the phone.''\em. Thus, we annotated them with the \em Request \em communicative function in the \em Task \em dimension.

\subsection{Instruct More Quiet}
In these turns the system asks the user speak more quietly using a polite request, such as \em ``I can't understand loud speech. Please speak more quietly.'' \em. Similarly to the turns annotated with the previous label, we annotated these with the \em Request \em communicative function in the \em Task \em dimension.

\subsection{Offer Help}
All turns annotated with this label contain the same utterance {--} \em ``To get help at any time, just say Help or press zero.'' \em {--} which instructs the user on how to get help. Accordingly, we annotated them with the \em Instruct \em communicative function in the \em Task \em dimension.

%
%

%
%

\section{Mapping of User Labels}
\label{sec:user}

Contrarily to system turns, the user turns are open and contain recognition errors. Thus, the mapping of their labels into the communicative functions of the standard is not as straightforward. Still, since the turns are typically short and the labels are highly domain-dependent, there are still mapping strategies that can be applied. For each label they are presented below.  

\subsection{Confirm}
\label{ssec:udaconfirm}
The turns annotated with this label consist of user confirmations that the system understood correctly, such as \em ``Correct'' \em and \em ``Yes''\em. Thus, we annotated them with the \em Confirm \em communicative function in the \em Task \em dimension. Furthermore, we also annotated them with the \em Allo Positive \em communicative function in the \em Allo-Feedback \em dimension, since they serve as feedback for the system.

\subsection{Confirm Bus}
The turns annotated with this label are in all similar to the ones annotated with the \em Confirm \em label (Section~\ref{ssec:udaconfirm}). They have a different label since they correspond to confirmations of the bus identifier specifically. Thus, we also annotated them with the \em Confirm \em and \em Allo Positive \em communicative functions in the \em Task \em and \em Allo-Feedback \em dimensions, respectively.   

\subsection{Confirm Departure}
The turns annotated with this label are in all similar to the ones annotated with the \em Confirm \em label (Section~\ref{ssec:udaconfirm}). They have a different label since they correspond to confirmations of the place of departure specifically. Thus, we also annotated them with the \em Confirm \em and \em Allo Positive \em communicative functions in the \em Task \em and \em Allo-Feedback \em dimensions, respectively.

\subsection{Confirm Destination}
The turns annotated with this label are in all similar to the ones annotated with the \em Confirm \em label (Section~\ref{ssec:udaconfirm}). They have a different label since they correspond to confirmations of the destination stop specifically. Thus, we also annotated them with the \em Confirm \em and \em Allo Positive \em communicative functions in the \em Task \em and \em Allo-Feedback \em dimensions, respectively.  

\subsection{Confirm Time}
The turns annotated with this label are in all similar to the ones annotated with the \em Confirm \em label (Section~\ref{ssec:udaconfirm}). They have a different label since they correspond to confirmations of time information specifically. Thus, we also annotated them with the \em Confirm \em and \em Allo Positive \em communicative functions in the \em Task \em and \em Allo-Feedback \em dimensions, respectively.  

\subsection{Goodbye}
Similarly to the system turns annotated with the \em Goodbye \em label (Section~\ref{ssec:sdagoodbye}, these turns intend to end the dialog by using the \em ``Goodbye'' \em keyword. Consequently we also annotated them with the \em Goodbye \em communicative function in the \em Social Obligations Management \em dimension.

\subsection{Inform}
There are only three turns annotated with this label and one of them should be annotated with the \em Time Information \em label (Section~\ref{ssec:udatimeinfo}) instead. The remaining two provide information unrelated to the system's question and were annotated with the \em Inform \em communicative function in the \em Task \em dimension.

\subsection{Line Information}
The turns annotated with this label provide information about the bus that the user wants to obtain information about. However, this information may come in the form of a question, such as \em ``When is the next 28X from Downtown to the Airport?''\em , or a statement, such as \em ``The 61A''\em. In the \em Task \em dimension, cases such as the first were annotated with the \em Set Question \em communicative function, while cases such as the second were annotated with \em Inform\em.

\subsection{Place Information}
The turns annotated with this label answer a system prompt for a departure or destination bus stop. Examples are \em ``Downtown'' \em and \em ``Duquesne''\em. Since these are answers to a specific question, in the \em Task \em dimension, we annotated them with the \em Answer \em communicative function instead of \em Inform\em. 

\subsection{Time Information}
\label{ssec:udatimeinfo}
Similarly to the previous, the turns annotated with this label answer a system prompt. In this case, related to time information. Examples are \em ``Eleven o'clock'' \em and \em ``Now''\em. Consequently, we also annotated them with the \em Answer \em communicative function in the \em Task \em dimension.

\subsection{New Query}
In these turns, the user instructs the system to start a new query, with utterances such as \em ``Start a new query''\em, structuring the dialog in that direction. Thus, on the one hand we annotated them with the \em Instruct \em communicative function in the \em Task \em dimension and, on the other hand, with the \em Interaction Structuring \em communicative function in the \em Discourse Structuring \em dimension.

\subsection{Polite}
In the turns annotated with this label, the user thanks the system for some reason. An example is \em ``Thank you'' \em. Thus, we annotated them with the \em Thanking \em communicative function in the \em Social Obligations Management \em dimension.

\subsection{Reject}
\label{ssec:udareject}
The turns annotated with this label consist of user rejections or corrections of the system's understanding, such as \em ``No'' \em and \em ``No, I need the next bus'' \em. Thus, we annotated them with the \em Disconfirm \em label in the \em Task \em dimension. Furthermore, we also annotated them with the \em Allo Negative \em communicative function in the \em Allo-Feedback \em dimension, since they serve as feedback for the system.

\subsection{Reject Bus}
The turns annotated with this label are in all similar to the ones annotated with the \em Reject \em label (Section~\ref{ssec:udareject}). They have a different label since they correspond to rejections of the bus identifier specifically. Thus, we also annotated them with the \em Disconfirm \em and \em Allo Negative \em communicative functions in the \em Task \em and \em Allo-Feedback \em dimensions, respectively.

\subsection{Reject Departure}
The turns annotated with this label are in all similar to the ones annotated with the \em Reject \em label (Section~\ref{ssec:udareject}). They have a different label since they correspond to rejections of the place of departure specifically. Thus, we also annotated them with the \em Disconfirm \em and \em Allo Negative \em communicative functions in the \em Task \em and \em Allo-Feedback \em dimensions, respectively.

\subsection{Reject Destination}
The turns annotated with this label are in all similar to the ones annotated with the \em Reject \em label (Section~\ref{ssec:udareject}). They have a different label since they correspond to rejections of the destination stop specifically. Thus, we also annotated them with the \em Disconfirm \em and \em Allo Negative \em communicative functions in the \em Task \em and \em Allo-Feedback \em dimensions, respectively.

\subsection{Reject Time}
The turns annotated with this label are in all similar to the ones annotated with the \em Reject \em label (Section~\ref{ssec:udareject}). They have a different label since they correspond to rejections of time information specifically. Thus, we also annotated them with the \em Disconfirm \em and \em Allo Negative \em communicative functions in the \em Task \em and \em Allo-Feedback \em dimensions, respectively.

\subsection{Request Help}
In the turns annotated with this label, the user asks for help using the keyword \em ``Help'' \em or by pressing the corresponding numeric key. Thus, we annotated them with the \em Request \em communicative function in the \em Task \em dimension.

\subsection{Request Next Bus}
The turns annotated with this label consist of the user asking for information about the next bus. However, this request may come in the form of a question, such as \em``When is the next bus?''\em, or a statement, such as \em``The next bus''\em. Furthermore, when a statement is used, it may be in response to a time request by the system. Thus, in the \em Task \em dimension, we used three different communicative functions to annotate these turns according to their nature. Respectively, \em Set Question\em, \em Inform\em, and \em Answer\em.

\subsection{Request Previous Bus}
These turns are similar to the ones annotated with the previous label, but asking for information about the previous bus instead of the next. Thus, in a similar manner, we annotated them with the \em Set Question\em, \em Inform\em, or \em Answer \em communicative functions in the \em Task \em dimension according to their different natures. 

\subsection{Request Schedule}
These turns consist of the user stating that he or she wants schedule information. An example is \em ``Holiday schedule''\em. Thus, we annotated them with the \em Inform \em communicative function in the \em Task \em dimension.

\subsection{Unqualified / Unrecognized}
The turns annotated with this label typically correspond to problems in the dialog. For instance, cases when the \ac{ASR} system failed to recognize most of the sentence and only outputted a small part of it, such as \em ``The''\em; cases of self-talk or third-party talk, such as \em ``I'm having fun''\em; and cases when the user says utterances unrelated to the task or that disrupt the dialog flow. Many of these turns are gibberish and do not correspond to any communicative function of the standard. However, some of them actually reveal an intention and, thus, should be annotated regardless of whether they make sense according to the flow of the dialog. Thus, these cases were annotated manually and individually with different labels.

%
%

%
%

\section{Discussion}
\label{sec:discussion}

\begin{table}[htbp]
    \caption{ISO 24617-2 communicative function distribution in the LEGO corpus.}
    \label{tab:legodistr}
    \centering
    \begin{tabular}{l l r r r r r r}
        \hline
        &                       & \multicolumn{2}{c}{\textbf{System}} & \multicolumn{2}{c}{\textbf{User}} & \multicolumn{2}{c}{\textbf{All}} \tabularnewline
        \textbf{Dimension}      & \textbf{Function} & \textbf{Count} & \textbf{\%} & \textbf{Count} & \textbf{\%} & \textbf{Count} & \textbf{\%}   \tabularnewline
        \hline
                                & Answer                    &     0 &  0.00              &  1462 & 28.70            &  1462 & 10.31             \tabularnewline
                                & Check Question            &  2256 & 24.84              &     1 &  0.02            &  2257 & 15.92             \tabularnewline
                                & Confirm                   &     0 &  0.00              &  1162 & 22.81            &  1162 &  8.20             \tabularnewline
                                & Disconfirm                &     0 &  0.00              &  1105 & 21.69            &  1105 &  7.79             \tabularnewline
        Task                    & Inform                    &   656 &  7.22              &   600 & 11.78            &  1256 &  8.86             \tabularnewline
                                & Instruct                  &  1812 & 19.95              &   106 &  2.08            &  1918 & 13.53             \tabularnewline
                                & Promise                   &   277 &  3.05              &     0 &  0.00            &   277 &  1.95             \tabularnewline
                                & Request                   &    70 &  0.77              &    85 &  1.67            &   155 &  1.09             \tabularnewline
                                & Set Question              &  1987 & 21.88              &   210 &  4.12            &  2197 & 15.50             \tabularnewline
                                & Suggest                   &    40 &  0.44              &     0 &  0.00            &    40 &  0.28             \tabularnewline
                                & \textbf{Total}            &  7098 & 78.15              &  4731 & 92.87            & 11829 & 83.44             \tabularnewline    
        \hline
                                & Allo Negative             &     0 &  0.00              &  1105 & 21.69            &  1105 &  7.79             \tabularnewline
        Allo-Feedback           & Allo Positive             &     0 &  0.00              &  1162 & 22.81            &  1162 &  8.20             \tabularnewline
                                & \textbf{Total}            &     0 &  0.00              &  2267 & 44.50            &  2267 & 15.99             \tabularnewline 
        \hline
                                & Auto Negative             &     0 &  0.00              &    19 &  0.37            &    19 &  0.13             \tabularnewline
        Auto-Feedback           & Auto Positive             &  3814 & 41.99              &    71 &  1.39            &  3885 & 27.40             \tabularnewline
                                & \textbf{Total}            &  3814 & 41.99              &    90 &  1.77            &  3904 & 27.54             \tabularnewline 
        \hline
        Discourse               & Interaction Structuring   &   852 &  9.38              &   103 &  2.02            &   955 &  6.74             \tabularnewline
        Structuring             & Opening                   &   347 &  3.82              &     0 &  0.00            &   347 &  2.45             \tabularnewline
                                & \textbf{Total}            &  1199 & 13.20              &   103 &  2.02            &  1302 &  9.18             \tabularnewline 
        \hline
                                & Apology                   &   163 &  1.79              &     3 &  0.06            &   166 &  1.17             \tabularnewline
        Social Obligations      & Goodbye                   &    36 &  0.40              &    31 &  0.61            &    67 &  0.47             \tabularnewline
        Management              & Greeting                  &   347 &  3.82              &    47 &  0.92            &   394 &  2.78             \tabularnewline
                                & Thanking                  &     0 &  0.00              &    10 &  0.20            &    10 &  0.07             \tabularnewline
                                & \textbf{Total}            &   546 &  6.01              &    91 &  1.79            &   637 &  4.49             \tabularnewline 
        \hline
        Time Management         & Pausing                   &   364 &  4.01              &     0 &  0.00            &   364 &  2.57             \tabularnewline
                                & \textbf{Total}            &   364 &  4.01              &     0 &  0.00            &   364 &  2.57             \tabularnewline 
    \end{tabular}
\end{table}

The result of applying the mapping process to the LEGO corpus is presented in Table~\ref{tab:legodistr}. We can see that communicative functions in the \em Task \em dimension are more predominant for user segments than for system segments, with 92.87\% of the user segments having communicative functions in that dimension versus the 78.15\% for system segments. While for system segments the most frequent functions are \em Check Question \em (24.84\%), \em Set Question \em (21.88\%), and \em Instruct \em (19.95\%), for user segments those are replaced by the \em Answer \em (28.70\%), \em Confirm \em (22.81\%), and \em Disconfirm \em (21.69\%) functions. This is coherent with the nature of the dialogs, which typically consists of the system questioning the user to obtain the required information. All segments considered, 83.44\% have communicative functions in the \em Task \em dimension. Since there are more system segments, the most frequent functions are the same as when considering those segments only, but with lower impact {--} 15.92\%, 15.50\%, and 13.53\%, respectively.

In terms of the feedback dimensions, 44.50\% of the user segments have functions in the \em Allo-Feedback \em dimension and no functions in the \em Auto-Feedback \em dimension. For system segments, the values are reversed, with 41.99\% of the segments having functions in the \em Auto-Feedback \em dimension and only 1.77\% in the \em Allo-Feedback \em dimension. Once again, these values are coherent with the nature of the dialogs, since the system typically uses feedback functions to check whether it understood what the user said, while the user confirms or disconfirms that.

As for the remaining semantic dimensions, it is important to refer that it is difficult to find communicative functions in those dimensions by simply converting the original labels of the LEGO corpus. Thus, the identified functions are just a small part of all those that exist in the corpus and that could be found through an manual and exhaustive processing of each segment. Still, it is interesting to notice that at least 13.20\% of the system segments contain communicative functions in the \em Discourse Structuring \em dimension. This reveals the rigid nature of the system's utterances and its intention to structure the dialog according to a specific path.        

Overall, by applying the mapping process described in this paper, 347 additional dialogs are annotated with the communicative functions of the ISO 24617-2 standard. Although communicative functions alone do not form a complete dialog act annotation according to the standard, the amount of generated data is important for dialog act research, especially for dialog act recognition experiments. This is particularly important given the reduced amount of existing data annotated according to the standard, which limits the conclusions that can be drawn from experiments using it.

%
%

%
%

\section*{Acknowledgements}
\label{sec:acknowledgements}

This work was supported by national funds through \ac{FCT} with reference UID/CEC/50021/2013, by Universidade de Lisboa, and by EU-IST FP7 project SpeDial under contract number 611396.

%
%

\bibliographystyle{template/splncs03}

\bibliography{references}

\end{document}